\pdfoutput=1

\documentclass[11pt]{article}

\usepackage{acl}

\usepackage{times}
\usepackage{latexsym}
\usepackage{multirow}   
\usepackage{graphicx}   
\usepackage{enumitem}
\usepackage{amsmath}    
\usepackage{amssymb}    
\usepackage{latexsym}
\usepackage[inkscapelatex=false]{svg}
\usepackage{url}    
\usepackage{colortbl, booktabs} 
\usepackage{makecell} 
\usepackage{color}  
\usepackage{array}
\usepackage{hyperref}   
\usepackage[all]{nowidow}   
\usepackage[subtle]{savetrees} 

\usepackage{algorithm}
\usepackage{algorithmic}
\usepackage{enumitem}
\usepackage{microtype}
\usepackage{multicol}
\usepackage{tipa}
\usepackage{arydshln} 
\usepackage{tipa}
\usepackage{tabularx}
\usepackage{booktabs}
\usepackage{inconsolata}

\usepackage[T1]{fontenc}

\usepackage[utf8]{inputenc}

\title{Seeing isn't Hearing:\\ Benchmarking Vision Language Models at Interpreting Spectrograms}

\author{Tyler Loakman\textsuperscript{1}, Joseph James\textsuperscript{1}, ~and Chenghua Lin\textsuperscript{2}\Thanks{Corresponding author}~~\\
  \textsuperscript{1}Department of Computer Science, University of Sheffield, UK \\
  \textsuperscript{2}Department of Computer Science, University of Manchester, UK \\
  \texttt{\{tcloakman1,jhfjames1\}@sheffield.ac.uk} \\
    \texttt{chenghua.lin@manchester.ac.uk} }

\begin{document}

\maketitle

\maketitle

\begin{abstract}
With the rise of Large Language Models (LLMs) and their vision-enabled counterparts (VLMs), numerous works have investigated their capabilities in tasks that fuse the modalities of vision and language. In this work, we benchmark the extent to which VLMs are able to act as highly-trained phoneticians, interpreting spectrograms and waveforms of speech. To do this, we synthesise a novel dataset containing 4k+ English words spoken in isolation alongside stylistically consistent spectrogram and waveform figures. We test the ability of VLMs to understand these representations of speech through a multiple-choice task whereby models must predict the correct phonemic or graphemic transcription of a spoken word when presented amongst 3 distractor transcriptions that have been selected based on their phonemic edit distance to the ground truth. We observe that both zero-shot and finetuned models rarely perform above chance, demonstrating the requirement for specific parametric knowledge of how to interpret such figures, rather than paired samples alone. 
\end{abstract}

\section{Introduction}

The ability of Large Language Models (LLMs) and Vision Language Models (VLMs) to reason about multimodal data has been studied extensively in recent years \cite{chia-etal-2024-puzzlevqa, lin-etal-2023-beneath, li-zhang-2023-cultural}. One of the most productive domains for testing the unification between vision and language is that of the explanation and creation of data visualisations, such as graphs and figures \cite{zhang-etal-2024-gpt, ge-etal-2024-automatic}. For the most part, such works have tested models' abilities to understand axes and labels, follow trendlines, and reformulate visual content as text (or vice versa). However, to date, little work has been performed on the ability of VLMs to understand more complex visualisations, where additional parametric knowledge must be combined with image inputs to reason about what is shown.\footnote{Code and resources are available at \url{https://github.com/tylerL404/seeing-is-hearing}.}

\begin{figure}
    \centering
    \includegraphics[width=1\linewidth]{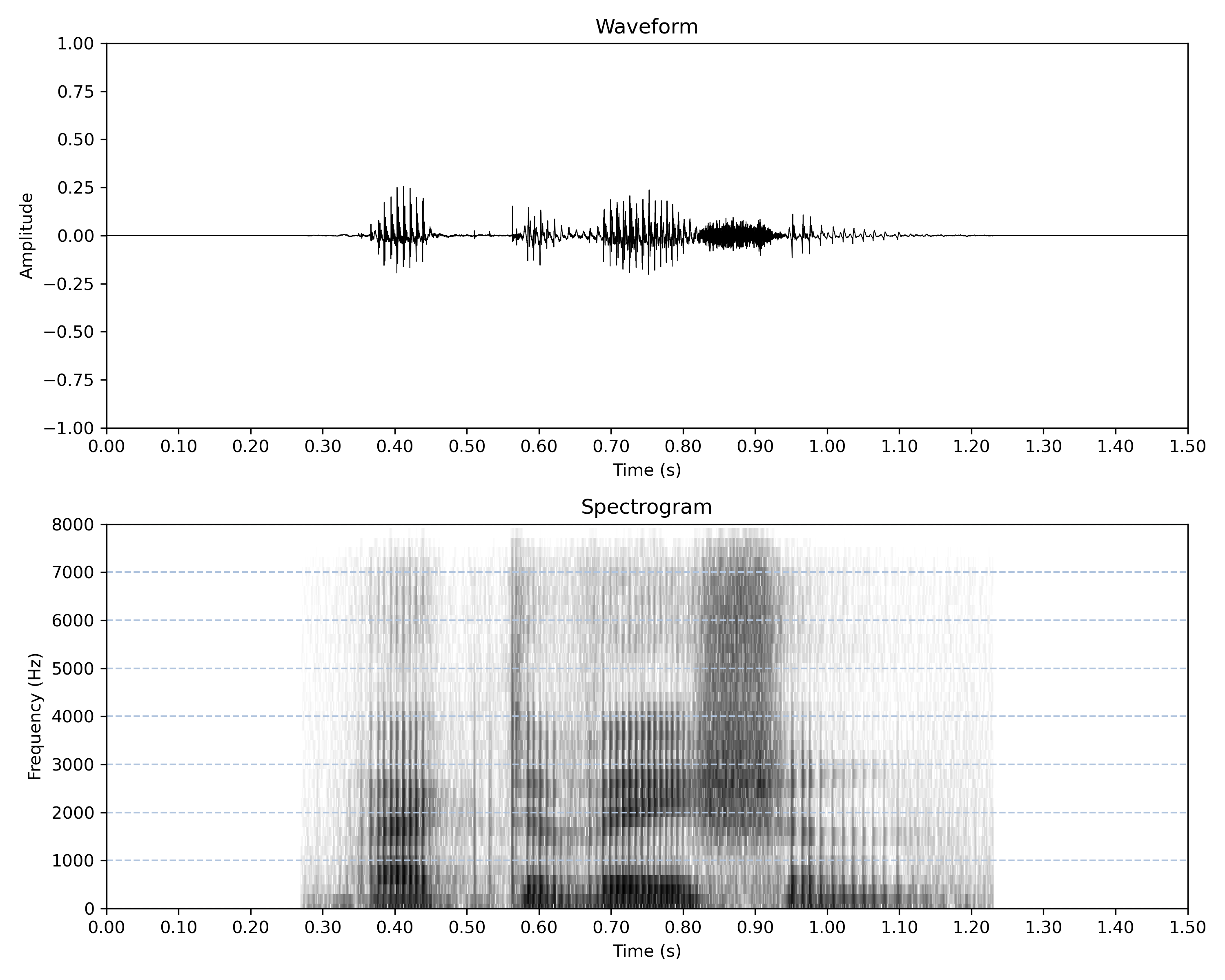}
    \caption{An example waveform (top) and spectrogram (bottom) of "activation" spoken by a text-to-speech model.}
    \label{fig:spectrogram_example}
\end{figure}

To that end, we present the task of spectrogram and waveform interpretation, whereby models must predict spoken words from visual representations of speech. As seen in \autoref{fig:spectrogram_example}, the x-axis of a spectrogram represents time, whilst the y-axis represents frequency, and the heat-map intensity represents the density of energy within a particular frequency range at a given time. On the other hand, a waveform also represents time on the x-axis, but displays the amplitude of a signal on the y-axis. Unlike more common visualisations (e.g., bar charts, pie charts, line graphs, etc.), in order to interpret such figures, models must reason about the visual content to establish what word is being spoken, using parametric knowledge regarding acoustic phonetics and the phonotactics of the target language to aid in reasoning about the observed signal characteristics such as vowel formants and fricative centre-of-gravity \cite{acoustic_phonetics_ladefoged}.

The contributions of our work are as follows:
\begin{itemize}
    \item We present a novel dataset of 4k+ words spoken in isolation using a range of synthetic voices, each with figures consisting of a spectrogram, a spectrogram with a waveform, and versions of the spectrograms/waveforms annotated with phone boundaries using forced-alignment.
    \item We perform what we believe to be the first analysis of whether VLMs are able to correctly interpret spectrograms and waveforms of speech in a phonologically-motivated multiple-choice paradigm.
\end{itemize}

\section{Related Work}
\label{sec:related_work}

Data-to-text tasks have long been a staple of Natural Language Processing (NLP) research \cite{castro-ferreira-etal-2019-neural}. Traditionally, such approaches have relied on access to largely text-based representations, such as converting tables to markdown or HTML \cite{he-etal-2023-anameta,liu-etal-2022-shot, eisenschlos-etal-2020-understanding}. However, the rise of VLMs has brought with it a range of benchmarks that concern image-based inputs \cite{zheng-etal-2024-multimodal} such as charts \cite{islam-etal-2024-large,zhou-etal-2023-enhanced} and graphs \cite{ai-etal-2024-advancement}, which have been fruitful in bearing benchmarks to assess the data-to-text abilities of VLMs \cite{zhu-etal-2025-benchmarking,islam-etal-2024-large,ai-etal-2024-advancement}. However, VLMs have been demonstrated to struggle with reasoning about inputs \cite{mukhopadhyay2024unraveling,hou2024vision} and easily being subjected to training biases \cite{vo2025vision}. Furthermore, recent works by \citet{loakman-etal-2024-ears} and \citet{alper2023kiki} have investigated the meta-level abilities of VLMs to reason about language, investigating whether or not they learn correlations between vision and text that can be used to simulate an understanding of sound symbolism.

Building upon these existing works in reasoning about visual data representations and exploring the meta-level understanding of VLMs about language, we present what we believe to be the first benchmark and analysis of the ability of VLMs to understand spectrograms and waveforms of speech, combining visual acuity with acoustic phonetic knowledge.

\section{Dataset Creation}
\label{sec:datset}

\subsection{Word List} 
\label{sec:word_list}
As a result of the complex effects of coarticulation in standard connected speech, we present VLMs with single English words spoken in isolation. To reduce ambiguity, our word list is required to be representative of standard dictionary entries rather than the acronyms, initialisms, and titles often found in word frequency lists. As a result, we use the Oxford 5000 word list,\footnote{\url{https://www.oxfordlearnersdictionaries.com/about/wordlists/oxford3000-5000}} consisting of words from A1-C1 difficulty on the Common European Framework of Reference for languages (CEFR) \cite{CouncilOfEurope2001CEFR}. We perform additional filtering to remove words that have more than one possible pronunciation in the CMU Pronouncing Dictionary, such as homographs, removing the need to manually check which version of the word our text-to-speech (TTS) system produces (owing to words being presented in isolation, void of contextual cues for the TTS model).\footnote{Accessed via the \textit{CMUDict} library: \url{https://pypi.org/project/cmudict/}} After filtering, this results in 4068 words.

\subsection{Speech Synthesis} 
\label{sec:speech_synthesis}
Owing to the high levels inter/intra-speaker variation present in human speech and the range of possible acoustic environments for recording \cite{JessenMichael2008FP}, we create a novel dataset using Microsoft's SpeechT5 trained for the task of TTS \cite{ao-etal-2022-speecht5} to speak each word in isolation. We use synthetic voices with a General American (GenAm) \cite{Wells1982} accent due to this enabling the use of the CMU Pronouncing Dictionary. A wider range of accents would require the manual creation of pronunciation dictionaries for each accent (intra-speaker variation notwithstanding), increasing the likelihood of phonetic realisations from the TTS model not accurately reflecting the transcriptions. We select 4 speaker embeddings from \textit{cmu-arctic-xvectors}\footnote{\url{https://huggingface.co/datasets/Matthijs/cmu-arctic-xvectors}} (2 male and 2 female perceived voices) for training, and 2 for testing (1 male and 1 female). We additionally add the General American voice from the Google Translate TTS API (via the gTTS library\footnote{\url{https://pypi.org/project/gTTS/}}) to the test set to mitigate biases from the characteristics of a particular TTS system. Further details are in Appendix \ref{apx:dataset-characteristics}.

\subsection{Spectrograms \& Waveforms}
\label{sec:figures}
As the image input to our VLMs, we provide a spectrogram of the target word being spoken by a TTS model, either on its own or with an accompanying waveform (akin to \autoref{fig:spectrogram_example}). To ensure consistency across our generated figures, we pad or truncate the audio files to 1.5 seconds in length for all words, which we validated to be longer than the speech portion of any of our generated audio files, ensuring no words have phonemes cut off. All spectrograms are generated with Librosa \cite{librosa}, using an \textit{nfft} of 128, a hop-length of 22, and a dynamic range of 70dB, using a Hann window. We find these settings to ensure an adequate frequency resolution without unnecessarily detailed harmonic information. See \autoref{fig:spectrogram_example} for an example of one of the generated figures.

As an additional feature of our dataset, we provide spectrograms and waveforms with vertical lines at approximate phone boundaries using the Montreal Forced Aligner \cite{mfa} with the respective US English dictionary and pre-trained acoustic model. Owing to resource constraints, annotated figures are not used in the following experiments.

\begin{figure*}[h!]
    \centering
    \includegraphics[width=0.85\linewidth]{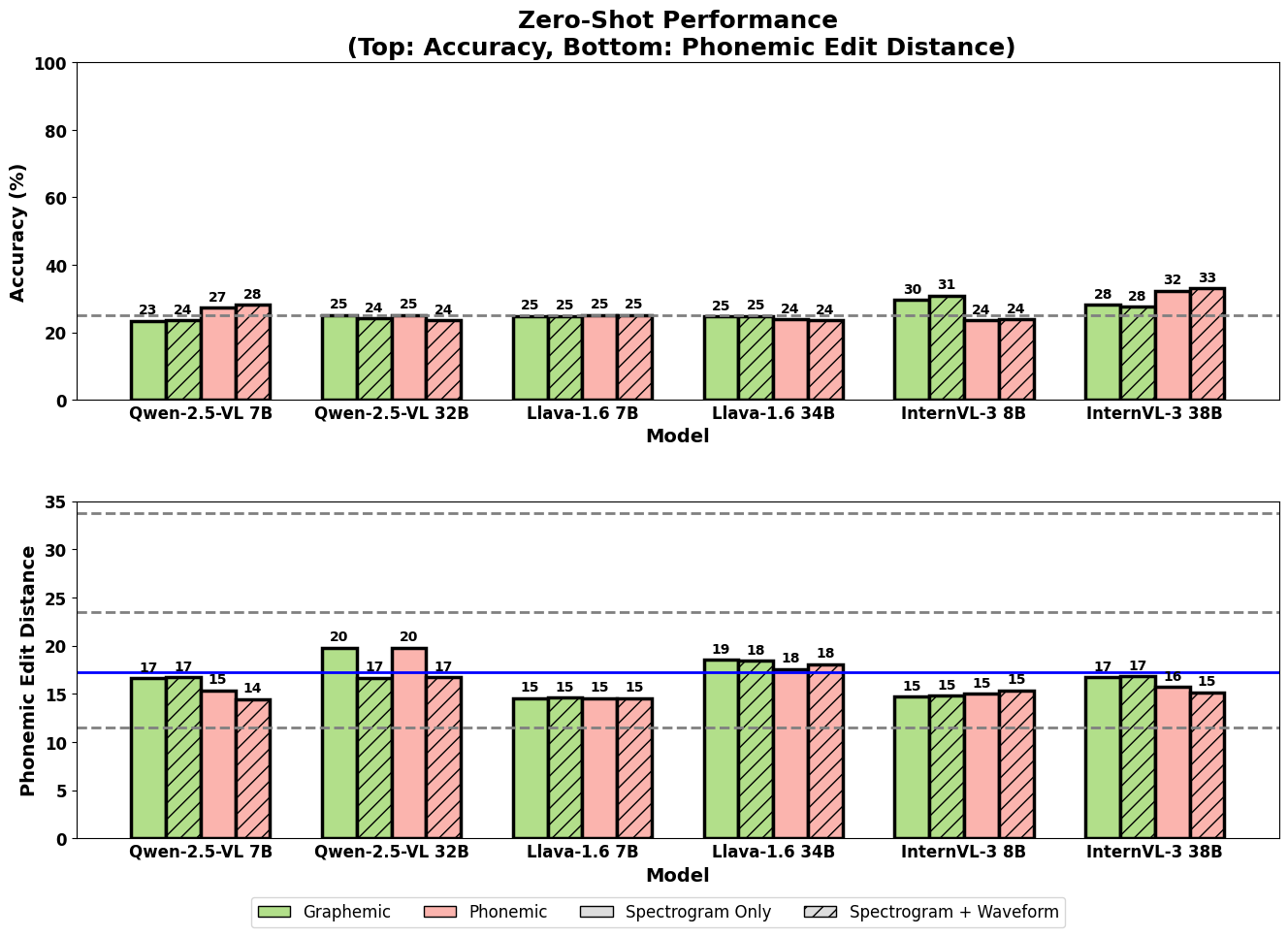}
    \caption{Zero-shot results in the multiple choice spectrogram interpretation task. \textit{Graphemic} refers to questions where the options were presented in their standard written English form, whilst \textit{Phonemic} refers to questions where the options were written in the International Phonetic Alphabet. \textit{Spectrogram}  and \textit{Spectrogram + Waveform }refer to the type of figure presented to the VLM. \textit{Accuracy} refers to the \% of the time that the correct answer was selected, whilst \textit{Phonemic Edit Distance} refers to the average distance of the selected option in comparison to the correct answer. The solid horizontal line in the Accuracy plot presents chance level agreement (25\%), whilst the dashed lines in the phonemic distance plot relate to the expected phonemic distance for consistently selecting the 2\textsuperscript{nd}, 3\textsuperscript{rd} or 4\textsuperscript{th} most similar option, whilst the solid blue line represents what is expected from random selection.}
    \label{fig:zero-shot}
\end{figure*}

\begin{figure}[h]
    \centering
    \includegraphics[width=1\linewidth]{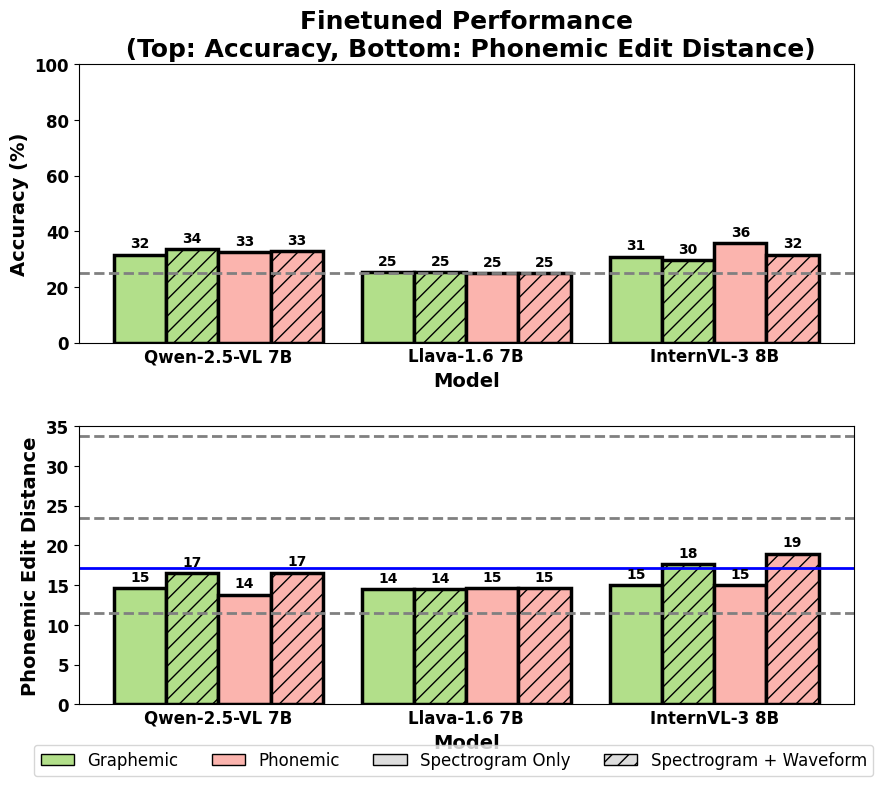}
    \caption{Finetuned results in the multiple choice spectrogram/waveform interpretation task. Please refer to \autoref{fig:zero-shot} for axis/condition information.}
    \label{fig:finetuned}
\end{figure}

\section{Methodology}
\label{sec:methodology}

\subsection{Task Setup}
\label{sec:tasks}
To test the ability of VLMs to interpret spectrograms, we use a multiple-choice question (MCQ) paradigm where the ground truth is paired with 3 distractors which are required to be carefully selected \cite{goel2025enhancing,alhazmi2024distractor,ding-etal-2024-learn}. Importantly, we select the distractors based on their phonemic edit distance (PED) from the ground truth using the \textit{weighted feature edit distance} function from the Panphon python package \cite{mortensen-etal-2016-panphon}, which is based on phonological distinctive features such as place and manner of articulation \cite{sound_patterns_of_english_chomsky}. This way, we select distractors that are phonemically similar or distant to the ground truth in order to identify if the VLMs are making informed choices (even if incorrect) or guessing randomly, owing to the semi-ambiguous nature of interpreting speech directly from a spectrogram. Specifically, the distractors consist of the word with the lowest and highest PED from the ground truth (i.e., the most and least similar, respectively) that have less than 50\% overlap in consecutive phonemes,\footnote{In early testing, we observed that models were able to learn to reach 50\% accuracy without being presented any input image. This was because the ground truth and the word with the lowest phonemic edit distance were frequently minimal pairs (e.g. "sheer" and "cheer"). It is for this reason that we introduce the overlap rule.} in addition to the word with the median PED. We additionally disallow distractors that start with the same character in their graphemic form to avoid overt biasing signals during training. The filtered list of 4068 words is broken into an 80-10-10 split for training, development, and testing, ensuring equal distribution of words based on length (in phonemes). See Appendix \ref{apx:dataset-characteristics} for further details on the split.

To mitigate ordering bias \cite{pezeshkpour-hruschka-2024-large, wei-etal-2024-unveiling}, we generated four permutations of each word for both the training and development splits. The target word appeared in each of the four possible positions (index 0 to 3), whilst the distractor positions were randomised.

\paragraph{Conditions} 
In total, we perform our tests under 4 conditions: graphemic  (i.e., standard written words) or phonemic (i.e., International Phonetic Alphabet) options, and using spectrograms alone or spectrograms with waveforms as the image input.

\subsection{Models}
\label{sec:models}
For this task, we selected a range of open-source VLMs from different model families. For zero-shot performance, we investigate the results from both small and large model variants, whilst for finetuning, we focus only on the small variants, owing to resource constraints. Specifically, we selected: Qwen 2.5-VL (7B and 32B) \cite{bai2025qwen25vltechnicalreport}, Llava 1.6 (7B and 34B) \cite{liu2024llavanext}, and InternVL 3 (8B and 38B) \cite{zhu2025internvl3exploringadvancedtraining}. See Appendix \ref{apx:implementation} for training hyperparameters, specific model names, and the prompt given to VLMs.

\section{Results}
\label{sec:results}

First, we present zero-shot performance results in \autoref{fig:zero-shot}. Regarding accuracy, we consistently observe that all models perform around chance-level (25\%) across all conditions. We observe no benefit from the additional information presented by the inclusion of the waveform (which would be useful for identifying speech sounds such as plosives), with performance even decreasing for Qwen 2.5-VL when waveforms are presented. Furthermore, regarding the phonemic and graphemic conditions, performance is generally higher when the options are presented as phonemes (Qwen 2.5-VL and InternVL-3 38B), but this does not hold for other models.

We hypothesise that the phonemic condition is more likely to activate knowledge directly related to acoustic phonetics, which would be relevant in the task of spectrogram/waveform understanding. On the other hand, regarding phonemic edit distance, the average distance of the selected option suggests frequent selection of low-distance distractors. Whilst a random selection is expected to have a phonemic distance of 17.2, most models perform slightly better than this, suggesting the tested models are making somewhat informed decisions, even if not selecting the correct answer.

Next, we present the results of finetuning the smaller model variants in \autoref{fig:finetuned}. We observe higher levels of performance than the zero-shot condition (particularly for InternVL-3). However, in our testing, we do not find a statistically significant difference between the performance of the finetuned models when tested with and without image inputs, suggesting that this performance increase is the result of learning the remaining biases in the dataset (owing to more similar words necessarily having similar phonemes). This highlights the difficulty of the task of spectrogram interpretation.

\subsection{ASR \& Trained Phonetician Comparison}
As a traditional baseline for speech understanding, we present the results of an Automatic Speech Recognition (ASR) system on the test set, using the original audio from the TTS systems. Specifically, we use Nvidia's \textit{canary-qwen-2.5b} model, which currently ranks among the top-performing systems on the Open ASR leaderboard (i.e., lowest word error rate).\footnote{\url{https://huggingface.co/spaces/hf-audio/open_asr_leaderboard}} When accounting for casefolding and punctuation removal to determine exact matches, the ASR model achieves an accuracy of 87.56\%, demonstrating that the underlying task is easily solvable when the acoustic information is given as input to purpose-built ASR system. This contrasts with the performance of the VLMs, which struggled when the same information was presented visually as spectrograms and waveforms.

Finally, we conducted human evaluation to establish an upper bound for the task of spectrogram/waveform interpretation using a trained phonetician, who is an author of this work. On a subset of 100 random examples from the test set on images consisting of a spectrogram and waveform (without boundary annotations). Importantly, a different author was responsible for generating the test set and presenting it, in order to avoid bias from pre-exposure.  We observe an accuracy of 75.00\% and an average phonetic edit distance of 5.17, highlighting the considerable gap between human and model performance, whilst demonstrating that there is sufficient information within the visualisations to achieve high accuracy.\footnote{We expect layperson accuracy on this task to be no better than chance.}

\section{Conclusion}
\label{sec:conclusion}

In this work, we presented the first analysis of the ability of VLMs to interpret speech from spectrograms and waveforms, using phonetically and phonologically motivated multiple-choice question approach. We observed that both zero-shot and finetuned models struggle to identify the correct answer, demonstrating the difficulty of the task and its potential as a benchmark assessment of the ability of VLMs to combine esoteric parametric knowledge with vision and language inputs.

\section*{Limitations}
\label{sec:limitations}

Owing to computational limitations, we were not able to finetune the larger model variants presented in this work or benchmark large vision-enabled reasoning models such as OpenAI's o1 and o3.
Furthermore, our dataset consists of synthetic speech to reduce the effects of co-articulation and acoustic environments. However, synthetic speech is less variable than natural human speech and therefore presents an easier form of the task.
Furthermore, we analyse models using zero-shot and finetuned approaches with a simple multiple-choice paradigm. However, we do not include any explicit knowledge within the prompt itself, which we leave to future work. We believe that future work may be best served by training models to recognise the characteristics of individual phonemes before learning entire words.


\subsubsection*{Acknowledgments}
Tyler Loakman and Joseph James are supported by the Centre for Doctoral Training in Speech and Language Technologies (SLT) and their Applications funded by UK Research and Innovation [grant number EP/S023062/1].

\bibliography{custom}

\appendix
\section{Appendix}

\subsection{Dataset Characteristics}
\label{apx:dataset-characteristics}

\begin{figure*}
    \centering
    \includegraphics[width=1\linewidth]{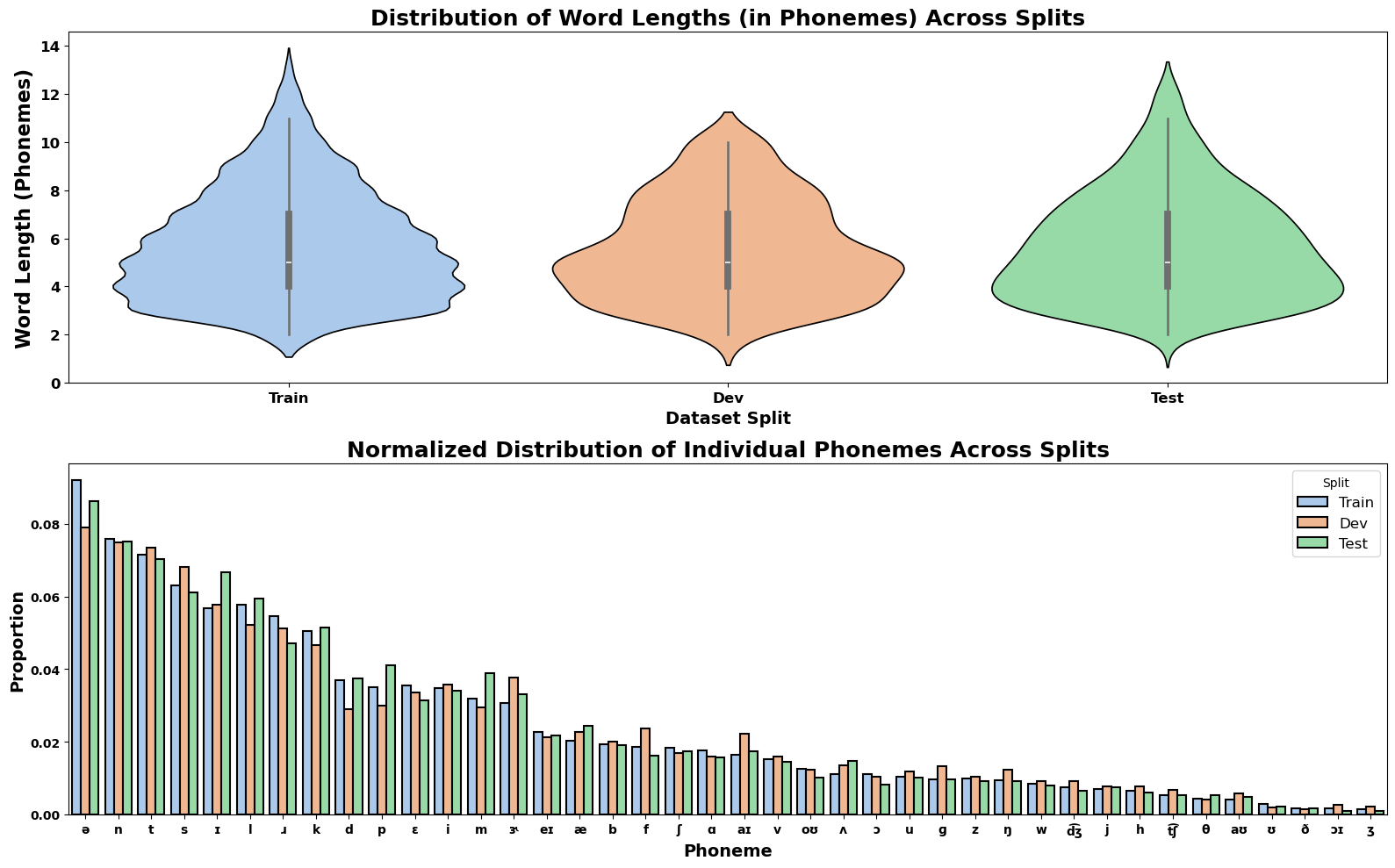}
    \caption{Distribution of word lengths (as determined via phoneme count) and individual phonemes across the training, development and test sets for finetuned VLMs.}
    \label{fig:dataset-distribution}
\end{figure*}

\autoref{fig:dataset-distribution} presents the distribution of words by length (in phonemes) across the training, development and test sets, in addition to the distribution of specific phonemes.

\subsection{Implementation Details}
\label{apx:implementation}

\paragraph{TTS Voices}
From \textit{cmu-arctic-xvectors}, we use speaker embeddings 69 (M), 750 (M), 2500 (F), and 3333 (F) for training, and 4444 (M) / 7500 (F) for testing. We additionally use Google Translate's General American TTS voice (which is only available in a female-presenting voice) for testing.

\paragraph{Model Selection}
All of our selected models were taken from Hugging Face. Specifically, for Qwen 2.5-VL we used \textsc{Qwen/Qwen2.5-VL-7B-Instruct} and \textsc{Qwen/Qwen2.5-VL-32B-Instruct}, for Llava 1.6 (i.e., Llava NeXT) we used \textsc{llava-hf/llava-v1.6-vicuna-7b-hf} and \textsc{llava-hf/llava-v1.6-34b-hf}, and for InternVL 3 we used \textsc{OpenGVLab/InternVL3-8B-hf} and \textsc{OpenGVLab/InternVL3-38B-hf}. For our ASR model we used \textsc{nvidia/canary-qwen-2.5b}.

\paragraph{Model Prompt}

\texttt{You are a Vision-Language Model specialized in phonetic interpretation of speech spectrograms. Your primary role is to act as a highly trained phonetician: given visual representations of spoken English words in the General American accent (spectrograms and/or waveforms), you must determine which graphemic or phonetic transcription correctly matches the spoken word.}

\texttt{For each example, you will be shown:}

\texttt{A spectrogram (and optionally a waveform) of a single English word spoken in isolation.}

\texttt{Four candidate transcriptions labeled 0 to 3 (one correct, three phonetic distractors).}

\texttt{Provide the final label as <label>label</label>, where label is the number (0-3) of the correct transcription.}

\texttt{Just output the label and nothing else in the format: <label>label</label>}

\paragraph{Training Details} We finetuned and performed inference for all models on a single A100 GPU. Hyperparameters used during training are presented in \autoref{tab:training-details}.

\begin{table}[ht]
\centering
\begin{tabularx}{0.85\linewidth}{l l}
\toprule
\textbf{Hyperparameter}           & \textbf{Setting} \\
\midrule
Epochs                            & 5 \\
Batch Size                       & 1 \\
Gradient Accumulation Steps     & 8 \\
Gradient Checkpointing           & True \\
Optimiser                             & AdamW \\
Learning Rate (LR)                    & 2e-4 \\
LR Scheduler Type               & constant \\
Weight Decay                     & 0.01 \\
Maximum Gradient Norm                   & 1.0 \\
Warmup Steps                     & 150 \\
Logging Steps                    & 150 \\
Evaluation Steps                       & 150 \\
Evaluation Strategy                    & steps \\
Save Steps                       & 150 \\
Early Stopping Callback             & 3 \\
\bottomrule
\end{tabularx}
\caption{Training hyperparameters.}
\label{tab:training-details}
\end{table}

\section{Statistical Testing}
We conducted Chi-squared tests for each condition to determine whether it had an effect on the output. The test aggregated all models and compared the effect of the word type (graphemic v. phonemic) and input type (spectrogram v. spectrogram and waveform). We found no significant effects of any conditions at $\alpha$ = 0.05.

\end{document}